\documentclass[runningheads]{llncs}

\usepackage{eccv}
\usepackage{eccvabbrv}
\usepackage{graphicx}
\usepackage{booktabs}
\usepackage{amsmath,amssymb}
\usepackage{hyperref}
\usepackage{placeins}
\usepackage{needspace}
\usepackage{float}

\begin{document}

\title{Small-Pollinator Detection in Cluttered Field Video}
\titlerunning{Small-Pollinator Detection in Cluttered Field Video}

\author{Onur Onal\inst{1} \and Chen Chen\inst{2}}
\authorrunning{O. Onal and C. Chen}

\institute{
Iowa State University, Ames, IA, USA\\
\email{onur@iastate.edu}
\and
Institute of AI, UCF\\
\email{chen.chen@ucf.edu}
}

\maketitle

\begin{abstract}
Detecting pollinators in field video is challenging: targets are small,
visually similar, and observed against cluttered vegetation under blur
and occlusion. We present a systematic empirical study of
small-pollinator detection under a practical single-GPU compute budget.
Using the BuzzSpot challenge dataset, we compare YOLO and RF-DETR models
across input resolutions and evaluate sliced inference, class-gated
fusion, size-routed ensembling, and post-hoc temporal processing.
RF-DETR Large at 1344-pixel resolution achieved our best hidden-test result, reaching 0.405 $\mathrm{mAP}_{50:95}$ and outperforming the 1120-pixel model (0.379) and the best single-model YOLO26m baseline (0.366).
The strongest gains came from adopting RF-DETR and increasing its input resolution, indicating that detector choice and input resolution were more effective levers than added inference-time complexity; the resolution gain was strongest for small objects and the rarer bumblebee and moth classes.
Sliced-inference fusion, size-routed ensembling, and warm-started
1536-pixel continuation did not surpass this result, while post-hoc
temporal processing did not improve the leaked diagnostic evaluation.
Error analysis identified bee--hoverfly discrimination as the clearest
remaining bottleneck: neighboring frames rarely supplied correctly classified hoverfly evidence for post-hoc correction. These
findings motivate learned feature-level temporal aggregation before the
final classification decision.

Code, experiment notebooks, and additional implementation details are
available at \url{https://github.com/OnurOnal7/buzzspot-pollinator-detection}.

\end{abstract}

\section{Introduction}
\label{sec:introduction}

Automated pollinator monitoring can support ecological analysis by
providing repeatable measurements of pollinator activity across field
conditions. In practice, however, pollinators occupy only a small
fraction of a high-resolution frame and are observed against cluttered
vegetation, often under motion blur, partial occlusion, and changing
illumination. The BuzzSpot challenge~\cite{buzzspot2026,buzzspot_codabench2026} reflects these conditions through
1920$\times$1080 field-video frames containing bees, bumblebees,
hoverflies, and moths, with five preceding frames provided for each
annotated keyframe.

This setting raises several practical questions. Higher input
resolution may preserve information needed for very small targets, but
it substantially increases memory and training cost. Image slicing can
magnify small objects, yet it may also remove useful scene context and
increase background false positives. Model ensembles can combine
complementary strengths, but routing and fusion introduce additional
failure modes. Finally, neighboring frames appear to offer a natural
source of temporal evidence, although post-hoc association can only
help when earlier predictions contain information that is more reliable
than the current-frame output.

We conduct a systematic empirical study of these trade-offs in a practical single-GPU setting, where higher resolution, larger models, and broader experimentation compete for limited compute. Starting from YOLO-based baselines, we
evaluate RF-DETR~\cite{rfdetr}, increased input resolution,
slicing-assisted inference, class-gated fusion, a size-routed
multi-model ensemble, and causal post-hoc temporal association. Final
system comparisons use the hidden-test evaluation server. After merging
the annotated training and validation splits, local validation scores
are treated only as contaminated diagnostic measurements and are kept
separate from hidden-test evidence.

Our strongest submitted system was a single RF-DETR Large model trained
at 1344-pixel resolution, which reached 0.405
$\mathrm{mAP}_{50:95}$ on the hidden test set. RF-DETR at 1120
resolution also outperformed the strongest submitted YOLO
configuration, while increasing RF-DETR resolution from 1120 to 1344
provided a further improvement. In contrast, the tested
full-frame/sliced fusion, size-routed ensemble, and warm-started
1536-pixel continuation did not surpass the 1344 model. Post-hoc temporal processing likewise did not improve the available local diagnostic evaluation.

Our main contributions are:

\begin{itemize}
    \item A compute-constrained empirical comparison of YOLO and
    RF-DETR configurations for small-pollinator detection, including
    hidden-test comparisons across detector families and RF-DETR input resolutions.

    \item An evaluation of slicing, class-gated fusion, size-based model
    routing, and post-hoc temporal association, showing that these
    additional mechanisms did not consistently improve generalization
    over the strongest single model.

    \item A temporal error analysis identifying bee--hoverfly
    discrimination as the clearest remaining bottleneck. Neighboring
    frames rarely contained correctly classified hoverfly evidence,
    limiting the corrective value of post-hoc temporal processing and
    motivating learned feature-level temporal aggregation.
\end{itemize}

\section{Related Work}
\label{sec:related-work}

Computer vision has been used to automate pollinator counting, tracking,
and behavioral analysis in agricultural environments. Ratnayake
et al.~\cite{ratnayake2022pollination} developed a multi-camera system
for monitoring several insect groups in commercial berry production,
while Alex et al.~\cite{alex2024bees} studied YOLO-based bee detection
and automated reporting from video. These works demonstrate the
feasibility of vision-based pollinator monitoring. Our study instead
focuses on the relative behavior of detector architectures, resolution,
slicing, model fusion, and temporal post-processing for four
pollinator categories under hidden-test evaluation.

Small-object detection methods commonly address the loss of spatial detail by
preserving higher-resolution representations or allocating high-resolution
computation selectively. HRDNet processes multi-resolution inputs with backbones
of different depths and fuses the resulting features~\cite{liu2020hrdnet}, while
QueryDet first identifies candidate small-object regions from lower-resolution
features and then sparsely computes higher-resolution features where
needed~\cite{yang2021querydet}. Slicing provides a model-agnostic alternative:
SAHI performs training and inference over overlapping image crops to increase the
effective scale of small targets~\cite{sahi}. Our main experiments evaluate how far general-purpose YOLO and RF-DETR
detectors can be improved through input-resolution scaling and sliced
inference under a constrained compute setting. We also report a separate
train-only QueryDet diagnostic in Section~\ref{sec:negative-results}.

DETR formulates detection as direct set prediction using bipartite
matching and transformer-based global reasoning~\cite{detr}. DINO
improves the DETR family through denoising training and improved query
initialization~\cite{dino}, while RF-DETR targets real-time end-to-end
detection~\cite{rfdetr}. Video-detection methods generally incorporate
temporal information before the final prediction stage. FGFA aligns
and aggregates neighboring-frame features using optical
flow~\cite{fgfa}; SELSA aggregates semantic information across a
sequence~\cite{selsa}; and TransVOD performs spatiotemporal aggregation
over features and object queries~\cite{transvod}. In contrast, our
temporal method operates only on independently produced boxes and
scores. Its negative result therefore applies to post-hoc association,
not to learned feature-level temporal modeling.

\section{Task and Experimental Setup}
\label{sec:setup}

\subsection{Dataset and Evaluation}

We study the BuzzSpot pollinator-detection challenge organized at
CVPPA 2026~\cite{buzzspot2026,buzzspot_codabench2026}. The dataset
contains 1920$\times$1080 RGB video frames recorded under real field
conditions across four farmland types---Phacelia, maize, pasture, and
mixed vegetation---on 11 recording days. The detection categories are
bee, bumblebee, hoverfly, and moth. Each annotated keyframe is
accompanied by its five immediately preceding frames, allowing temporal
context to be used without requiring full videos. Blur and occlusion
attributes are also provided, although we did not use them as training
targets.

The organizers describe approximately 15,000 annotated instances across
the complete dataset. In the final release, the publicly annotated
training and validation splits contain 5,275 and 932 keyframes,
respectively, with exactly 12,000 bounding boxes in total. Their class
distribution is shown in Table~\ref{tab:dataset}. Each keyframe forms a
six-frame sequence, giving 31,650 training frames and 5,592 validation
frames when the unannotated context frames are included. The extended
hidden test set contains 4,763 keyframes from 53 videos, together with
their context frames, but its bounding-box annotations were not
released.

\begin{table}[H]
\centering
\caption{Class distribution in the final annotated training and
validation splits}
\label{tab:dataset}
\setlength{\tabcolsep}{8pt}
\begin{tabular}{lrrr}
\hline
Class & Train & Validation & Combined \\
\hline
Bee        & 8,677 & 934 & 9,611 \\
Bumblebee  &   237 &  30 &   267 \\
Hoverfly   & 1,753 &  95 & 1,848 \\
Moth       &   217 &  57 &   274 \\
\hline
Total      & 10,884 & 1,116 & 12,000 \\
\hline
\end{tabular}
\end{table}

The primary leaderboard metric is mean average precision averaged over
IoU thresholds from 0.50 to 0.95, denoted
$\mathrm{mAP}_{50:95}$. The evaluation server additionally reports
$\mathrm{mAP}_{50}$ and class- and object-size-specific results. The challenge documentation states that predictions are associated with
ground-truth boxes using Hungarian assignment%
~\cite{buzzspot_codabench2026}. We therefore distinguish
official hidden-test scores from local COCO-style evaluation rather
than assuming that the two implementations are identical.

Our earliest YOLO baselines used the released training split while
holding out the validation split. For the final experiments, we merged
the annotated training and validation keyframes, as the supplied split
was not mandatory~\cite{buzzspot_codabench2026}. Consequently, evaluation on the original validation
images after this merge was training-contaminated. We refer to these
numbers as \emph{leaked validation} scores and use them only for
diagnostic comparisons, not as estimates of generalization. Comparisons
between final systems are based on the hidden-test leaderboard. The final phase permitted only ten total submissions and three
submissions per day~\cite{buzzspot_codabench2026}, limiting the number of configurations that could be evaluated
on unseen annotations.

\subsection{Models and Training}

All experiments were conducted in a high-memory Google Colab
environment using one NVIDIA L4 GPU with 22.5~GB of available GPU
memory, 53~GB of system memory, and approximately 236~GB of disk
storage.

We first established convolutional baselines using YOLO11s, YOLO26s,
and YOLO26m. We used YOLO26m for the final YOLO experiments because
it was the strongest YOLO configuration we tested while remaining
feasible on the single L4 GPU~\cite{yolo26}. Most final YOLO runs used an input size of 1536
pixels and retained mosaic augmentation. To reduce the effect of class
imbalance, we applied image-level oversampling with approximate
multipliers of $1\times$ for bee, $4\times$ for bumblebee, $2\times$
for hoverfly, and $5\times$ for moth. These multipliers were chosen heuristically from class frequency: bee
received no boost, hoverfly received a smaller boost, and the two rarest
classes, bumblebee and moth, received the largest boosts. They were not
tuned as optimal values. The principal full-frame
YOLO26m model was trained for 32 epochs on the combined annotated
training and validation data.

We then evaluated RF-DETR~\cite{rfdetr}. An eight-epoch RF-DETR Small
run at 1120-pixel resolution served as a feasibility experiment. This run showed enough potential to justify moving to RF-DETR Large to
test whether a higher-capacity model could improve accuracy within the
same hardware budget. The
main experiments used RF-DETR Large for 32 epochs at resolutions of
1120 and 1344 pixels while holding the remaining training configuration
fixed. Training used a batch size of two with eight-step gradient
accumulation, giving an effective batch size of 16. Exponential moving
average checkpointing and gradient checkpointing were enabled to make
training feasible on the single GPU. A further experiment initialized
from the 1344-pixel EMA checkpoint and continued at 1536 resolution for
12 epochs using a fresh optimizer and scheduler with a reduced learning
rate of $5\times10^{-5}$. The detector-family comparison was therefore a best-feasible-system
comparison under the available hardware budget, not a parameter- or
latency-matched architecture comparison. 

For RF-DETR inference, we evaluated confidence thresholds of 0.001,
0.005, 0.01, 0.02, 0.03, 0.05, 0.075, 0.10, 0.15, 0.20, and 0.30 on
the leaked former validation split, with predictions capped at 100 per
image. We selected 0.01 according to COCO-style
$\mathrm{mAP}_{50:95}$ and fixed this operating point before hidden-test
export.

\subsection{Alternative Inference Strategies}

\noindent\textbf{Slicing and class-gated fusion.}
We first applied slicing-assisted inference, motivated by the limited
pixel area occupied by many pollinators~\cite{sahi}. Directly applying
approximately $700\times700$ tiles with 20\% overlap to a model trained
on full frames produced many background false positives. We therefore
fine-tuned a second YOLO26m model on the sliced distribution, initialized
from the full-frame checkpoint. At inference, we evaluated both an
all-class fusion and a class-gated fusion. In the gated version, the
full-frame model contributed all four categories, whereas the sliced
model contributed only bee, bumblebee, and moth detections; its
hoverfly predictions were excluded before merging.

\noindent\textbf{Size-routed ensemble.}
We also constructed a three-model ensemble using the standard COCO
object-area ranges~\cite{coco}. Predicted boxes with area below
$32^2$ pixels were taken from the full-frame--sliced YOLO ensemble,
boxes between $32^2$ and $96^2$ pixels from RF-DETR Large, and boxes
above $96^2$ pixels from the full-frame YOLO model. The selected
detections were merged using class-aware non-maximum suppression and
limited to 100 predictions per image. Routing was based on predicted,
rather than ground-truth, box area.

\noindent\textbf{Post-hoc temporal association.}
For the temporal experiment, RF-DETR Large at 1344 resolution was run
independently on the keyframe and its five preceding frames.
Detections were associated backward through adjacent frames using
class-agnostic Hungarian assignment. Let $b_t$ be the most recent box
in a track and $b_c$ a candidate in the preceding frame. Their matching
cost was

\begin{align}
\mathcal{C}(b_t,b_c) ={}&
0.55\min\left(\frac{\lVert p_c-p_t\rVert_2}{D},8\right)
+0.20\min\left(\left|\log\frac{A_c}{A_t}\right|,1.8\right)
\nonumber\\
&+0.10\min\left(\left|\log\frac{r_c}{r_t}\right|,1.5\right)
+0.10\left(1-\operatorname{IoU}(b_c,b_t)\right)
\nonumber\\
&+0.05\min\left(\frac{\lVert p_c-\hat{p}_c\rVert_2}{D},8\right),
\label{eq:temporal-cost}
\end{align}

where $p$, $A$, and $r$ denote box center, area, and aspect ratio,
respectively; $D$ is the image diagonal; and $\hat{p}_c$ is a
constant-velocity center prediction obtained from the two most recent
track positions. Assignments with cost above 5.0 were rejected, and
tracks were not allowed to skip a frame. The association weights were selected heuristically to prioritize
normalized center displacement and scale consistency; we did not perform
a sensitivity analysis over these weights.

Tracking used detections with confidence above 0.001, capped at 200
proposals per frame; final outputs retained the 0.01 confidence
threshold and 100-detection limit. From the resulting trajectories, we evaluated score averaging, a bounded historical maximum, reliability-based reranking, recovery of low-confidence keyframe proposals, and a separate box-propagation ablation. We also implemented gated bee--hoverfly score fusion, but a pre-specified rescueability diagnostic disabled class correction before
candidate evaluation. These procedures modify only detector outputs and do not perform learned temporal feature aggregation.

\enlargethispage{5\baselineskip}
\section{Results}
\label{sec:results}

\subsection{Hidden-Test Performance}
\label{sec:hidden-results}

Table~\ref{tab:hidden-results} summarizes the systems submitted to the
hidden test set for which detailed server results are available. The
full-frame YOLO26m model reached 0.366
$\mathrm{mAP}_{50:95}$. Combining it with the sliced YOLO26m model
through class-gated fusion increased the score only slightly, to 0.368.
RF-DETR Large at 1120 resolution reached 0.379, improving on the
strongest submitted YOLO configuration by 0.011 mAP.

\begin{table}[H]
\centering
\caption{Hidden-test performance of submitted systems. Input lists the
inference resolution or resolutions used by each system. Results are
rounded to three decimals}
\label{tab:hidden-results}
\setlength{\tabcolsep}{4.2pt}
\begin{tabular}{lccc}
\hline
System & Input & $\mathrm{mAP}_{50:95}$ & $\mathrm{mAP}_{50}$ \\
\hline
YOLO26m, full frame
    & 1536 & 0.366 & 0.546 \\
YOLO26m, full+sliced gated
    & 1536+704 & 0.368 & 0.549 \\
RF-DETR Large
    & 1120 & 0.379 & 0.577 \\
Size-routed ensemble
    & 1536+704+1120 & 0.369 & 0.558 \\
RF-DETR Large
    & 1344 & \textbf{0.405} & \textbf{0.598} \\
RF-DETR Large, warm-started
    & 1536 & 0.395 & 0.590 \\
\hline
\end{tabular}
\end{table}

Increasing RF-DETR input resolution from 1120 to 1344 under otherwise
matched logged settings produced the strongest submission. The
1344-pixel model reached 0.405 mAP, improving over the 1120 model by
0.026 and over the strongest YOLO submission by 0.037. This comparison provides the clearest evidence in our experiments that higher input resolution improved hidden-test performance.

The improvement did not continue monotonically. Continuing from the
1344-pixel checkpoint at 1536 resolution for 12 additional epochs
reached 0.395 mAP, 0.010 below the 1344 model. Because this experiment
changed both resolution and training history, it does not establish
that 1536-pixel inputs are inherently worse. It shows that increasing
resolution through this warm-started continuation did not provide an
automatic improvement.

The size-routed ensemble also did not preserve the advantage of its
strongest component. It scored 0.369 mAP, below standalone RF-DETR at
both 1120 and 1344 resolution. Hard routing therefore added system
complexity without improving hidden-test generalization.

\subsection{Class and Object-Size Performance}
\label{sec:breakdown}

Table~\ref{tab:detailed-results} reports the class- and size-specific
hidden-test results. Following the evaluation server, size-specific AP
is reported at IoU 0.50, whereas class-specific AP is averaged over IoU
thresholds from 0.50 to 0.95.

\begin{table}[H]
\centering
\caption{Hidden-test AP by object-size category and pollinator class.
Size-specific AP is evaluated at IoU 0.50; class-specific AP is
averaged over IoU thresholds 0.50:0.95. The best value in each column
is shown in bold}
\label{tab:detailed-results}
\setlength{\tabcolsep}{3.2pt}
\resizebox{\textwidth}{!}{
\begin{tabular}{lccccccc}
\hline
System &
$AP_{\mathrm{S}}^{50}$ &
$AP_{\mathrm{M}}^{50}$ &
$AP_{\mathrm{L}}^{50}$ &
$AP_{\mathrm{bee}}^{50{:}95}$ &
$AP_{\mathrm{bum.}}^{50{:}95}$ &
$AP_{\mathrm{hov.}}^{50{:}95}$ &
$AP_{\mathrm{moth}}^{50{:}95}$ \\
\hline
YOLO26m, full frame
& 0.367 & 0.438 & 0.811
& 0.537 & 0.299 & 0.133 & 0.494 \\

YOLO26m, full+sliced gated
& 0.405 & 0.451 & 0.795
& \textbf{0.544} & 0.309 & 0.134 & 0.484 \\

RF-DETR, 1120
& 0.383 & \textbf{0.526} & 0.720
& 0.520 & 0.388 & 0.142 & 0.464 \\

Size-routed ensemble
& 0.389 & 0.516 & \textbf{0.812}
& 0.514 & 0.362 & 0.141 & 0.460 \\

RF-DETR, 1344
& \textbf{0.410} & 0.500 & 0.720
& 0.532 & \textbf{0.435} & \textbf{0.154} & \textbf{0.498} \\

RF-DETR, 1536 warm-start
& 0.398 & 0.522 & 0.715
& 0.531 & 0.419 & 0.141 & 0.491 \\
\hline
\end{tabular}
}
\end{table}

The class-gated full-frame/sliced YOLO ensemble had higher
small-object AP than the full-frame model, 0.405 compared with 0.367,
yet its overall mAP increased by only 0.002. Hoverfly AP was
effectively unchanged, while moth and large-object performance
decreased slightly. The result shows that the ensemble changed the
size-specific error profile without producing a substantial overall
improvement.

Relative to RF-DETR at 1120, the 1344 model improved AP for all four
classes: bee increased from 0.520 to 0.532, bumblebee from 0.388 to
0.435, hoverfly from 0.142 to 0.154, and moth from 0.464 to 0.498.
The size-specific changes were less uniform. Small-object AP increased
from 0.383 to 0.410, medium-object AP decreased from 0.526 to 0.500,
and large-object AP remained approximately unchanged. The resolution
gain therefore cannot be attributed uniformly to every object-size
range, although it improved small-object AP and all four class-specific
scores.

RF-DETR underperformed full-frame YOLO on large objects, reaching
0.720 compared with 0.811 $AP_{\mathrm{L}}^{50}$. This weakness
motivated routing large predicted boxes to YOLO in the size-routed
ensemble. The ensemble recovered a large-object AP of 0.812, but this
did not recover overall performance: its class-specific scores remained
below those of RF-DETR at 1344, and its overall mAP was only 0.369.
Strong performance in one size range was therefore insufficient to
offset losses across the remaining detections.

Hoverfly remained the weakest category by a substantial margin. Our best-performing model reached only 0.154 hoverfly AP, compared with 0.435 for
bumblebee and 0.498 for moth. This gap is notable because the training
set contains 1,753 hoverfly annotations but only 237 bumblebees and 217
moths. Annotation scarcity alone therefore does not explain the
hoverfly result. Section~\ref{sec:hoverfly-analysis} examines whether
neighboring frames contain enough correctly classified evidence to
repair these errors.

\subsection{Interventions That Did Not Transfer}
\label{sec:negative-results}

Several interventions produced either no local improvement or gains
that did not transfer to the hidden test set. Because most were
evaluated after merging the annotated training and validation splits,
their local scores are treated only as leaked diagnostic results.

Magnifying hoverflies through annotation-centered crop augmentation
created 3,506 additional training crops, but the resulting YOLO26s run
reached only 0.372 leaked-validation mAP, with hoverfly AP remaining
near 0.13. Hard-negative and tile mining was similarly inconclusive:
the mining procedure recovered 523 useful negatives rather than the
targeted 3,000, and the resulting model did not improve over the
corresponding full-frame baseline. Neither simple magnification nor the
available set of mined background examples was sufficient to resolve
the bee--hoverfly boundary.

Slicing showed a clearer difference between local and hidden-test
behavior. On the contaminated validation split, the full-frame
YOLO26m model reached 0.417 mAP, while the sliced model alone reached
0.345. Combining their predictions increased the local score to 0.431,
and excluding the sliced model's hoverfly predictions increased it
further to 0.434. On the hidden test set, however, class-gated
full-frame/sliced fusion increased mAP only from 0.366 to 0.368.
Slicing recovered some small detections, but additional false positives
and uneven class behavior limited its overall value.

The strongest mismatch between local and hidden evaluation occurred during the
warm-started 1536 experiment. Its leaked-validation score increased from
approximately 0.886 for the 1344 model to 0.921, while its hidden-test score
decreased from 0.405 to 0.395, reinforcing the need to treat post-merge
validation only as a diagnostic signal. More broadly, the interventions targeted
narrower failure modes than the persistent bee--hoverfly ambiguity: slicing and
crop augmentation addressed object scale, hard-negative mining targeted
background confusion, and size routing exploited size-specific model strengths.
The full-frame/sliced ensemble raised small-object AP@0.5 from 0.367 to 0.405
while overall hidden-test mAP increased only from 0.366 to 0.368. Similarly, the
size-routed ensemble recovered YOLO's stronger large-object AP@0.5
(0.812 vs.\ 0.720 for RF-DETR) yet reached only 0.369 overall mAP. Hoverfly AP
remained between 0.133 and 0.154 across submitted systems, suggesting limited
class-level complementarity for fusion or routing. Thus, the added mechanisms
mainly redistributed errors, while the single 1344-pixel RF-DETR model
generalized best.

As a separate train-only diagnostic, we selected
QueryDet~\cite{yang2021querydet} because its sparse query mechanism
concentrates high-resolution computation on regions likely to contain
small objects. This made it a direct test of whether a
small-object-specific architecture could address the scale-related
weaknesses observed in our earlier experiments. We trained its
ResNet-50-FPN configuration using the same image-level oversampling
scheme and a 1200-pixel short side, and evaluated it on the untouched
validation split. The run was stopped after validation had
plateaued. The best checkpoint reached 0.367
$\mathrm{mAP}_{50:95}$, but small-object AP remained only 0.079
$AP_{\mathrm{S}}^{50{:}95}$. Per-class AP was 0.501 for bee, 0.496 for
bumblebee, 0.171 for hoverfly, and 0.302 for moth. Hoverfly therefore
remained substantially weaker than bee and bumblebee despite having far more
training annotations than bumblebee. 

Earlier YOLO runs under the same
train-only and held-out-validation protocol achieved higher validation AP.
However, QueryDet was not retrained on the combined training and validation
splits or evaluated on the hidden test set, so this result is reported
separately and cannot be compared directly with the submitted systems. Within this diagnostic, the weak small-object AP and persistent hoverfly gap indicate that a small-object-oriented architecture alone was insufficient to resolve the persistent hoverfly weakness. Taken together with the RF-DETR error analysis in Section~\ref{sec:hoverfly-analysis}, where many hoverflies were well localized but classified as bees, this suggests that allocating additional high-resolution computation to small-object regions does not by itself address the remaining bee--hoverfly appearance ambiguity.

\section{Temporal Error Analysis and Discussion}
\label{sec:temporal-discussion}

\subsection{Temporal Post-Processing Results}
\label{sec:temporal-results}

We evaluated the temporal methods on the original validation split
using the 1344-pixel RF-DETR model. Because these images had already
been included in training, all results in this subsection are leaked
diagnostics rather than estimates of generalization. No temporal
candidate was submitted to the hidden test set.

The single-frame detector remained the strongest configuration across
the candidate sweep. Table~\ref{tab:temporal-results} compares it with
the highest-scoring non-baseline method, which increased each
keyframe score toward a bounded maximum from its associated historical
detections. The temporal candidate decreased COCO-style
$\mathrm{mAP}_{50:95}$ from 0.8866 to 0.8863. It also decreased mAP
from 0.8340 to 0.8255 under our custom Hungarian-style evaluator. The
latter is a local implementation based on the assignment procedure
described by the challenge organizers and should not be interpreted as
the private evaluation-server implementation.

\begin{table}[H]
\centering
\caption{Temporal post-processing on the leaked validation split. The
Hungarian-style metric is produced by our local evaluator. The temporal
row is the strongest non-baseline candidate under COCO-style
$\mathrm{mAP}_{50:95}$}
\label{tab:temporal-results}
\setlength{\tabcolsep}{5pt}
\begin{tabular}{lccc}
\hline
Method &
COCO $\mathrm{mAP}_{50:95}$ &
Hungarian $\mathrm{mAP}_{50:95}$ &
Detections \\
\hline
Single-frame baseline
& \textbf{0.8866} & \textbf{0.8340} & 17,637 \\
Bounded historical maximum
& 0.8863 & 0.8255 & 36,174 \\
\hline
\end{tabular}
\end{table}

The remaining variants included temporal score averaging,
reliability-based reranking over nine parameter settings, and a
separate box-propagation ablation. None exceeded the single-frame baseline. The bounded-maximum candidate also more than doubled the number of retained detections by promoting proposals from the low-threshold tracking pool, without improving precision sufficiently to raise mAP.

\subsection{Bee--Hoverfly Confusion and Temporal Evidence}
\label{sec:hoverfly-analysis}

We next examined hoverfly errors from the 1344-pixel RF-DETR model,
particularly confusion with bees, and whether the preceding frames
contained corrective evidence. The validation split contains 95 ground-truth hoverflies.
Under the diagnostic matching criteria, 79 were already detected
correctly in the keyframe, leaving 16 failures. Only one of these
failures met the criteria for rescue using correctly classified
detections from preceding frames, corresponding to a rescue rate of
6.25\%.

\Needspace{8\baselineskip}
The diagnostic also found 19 of the 95 hoverflies for which every
reliable associated historical proposal was classified as bee. These
proposals were drawn from the low-threshold tracking pool and were often
below the final output threshold. This indicates an absence of
corrective hoverfly evidence in neighboring frames, rather than
confident misclassification in every frame. In such cases, post-hoc
temporal adjustment has no correctly classified historical proposal to
promote.

Because the observed rescue rate fell below the pre-specified 8\% gate,
bee--hoverfly relabeling was disabled before candidate evaluation. This
avoided tuning a relabeling rule after observing that corrective
historical evidence was largely absent.

\begin{figure}[H]
\centering
\IfFileExists{hoverfly_confusion_grid.pdf}{
    \includegraphics[width=\linewidth]{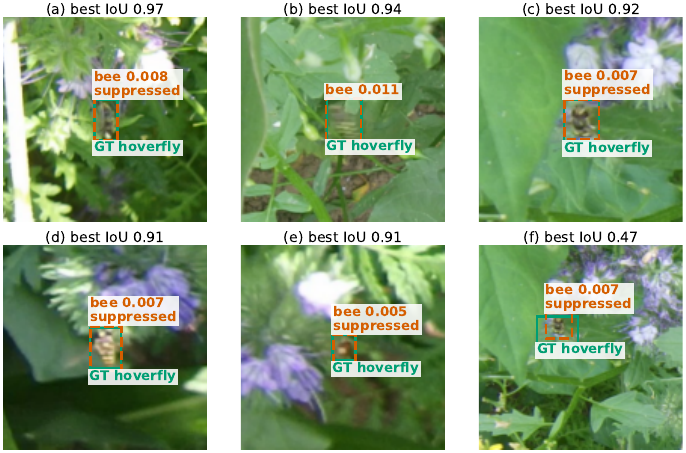}
}{
    \fbox{
        \parbox[c][0.22\textheight][c]{0.94\linewidth}{
            \centering
            Placeholder for a $2\times3$ grid of representative
            hoverfly failure cases from RF-DETR Large at 1344-pixel
            resolution.
        }
    }
}
\vspace{-1\baselineskip}
\caption{Selected keyframe hoverfly failures from RF-DETR Large at
1344-pixel resolution on the leaked validation diagnostic.
Ground-truth boxes and their highest-overlap proposals are shown with
predicted class, confidence, and IoU. Five of the six proposals have
IoU of at least 0.91 but are classified as bee, indicating incorrect
classification despite accurate localization. Predictions below the
final output threshold of 0.01 are marked as suppressed.}
\label{fig:hoverfly-confusions}

\vspace{0.5\baselineskip}
\includegraphics[width=\linewidth]{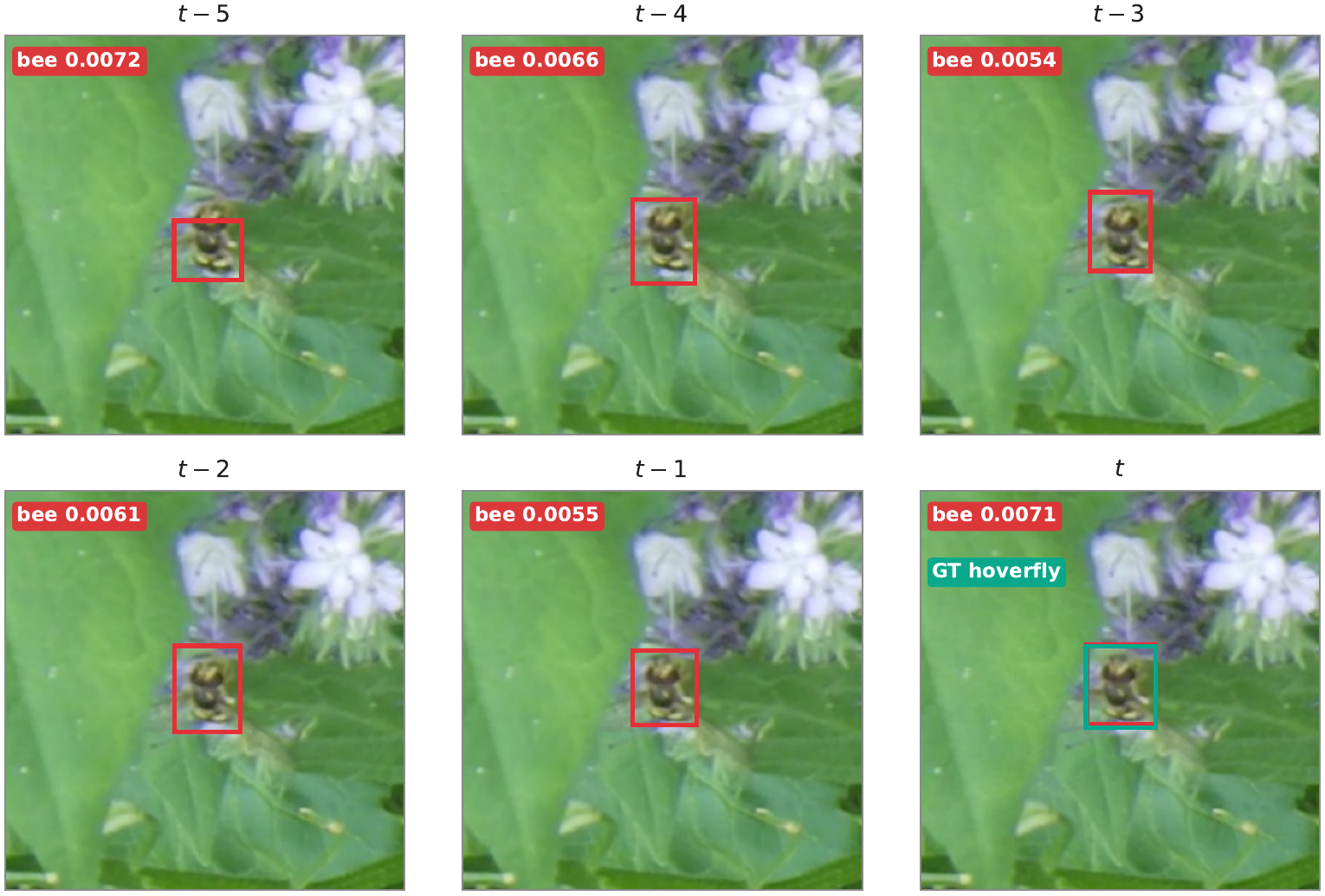}
\caption{Representative bee--hoverfly temporal failure case from the
leaked validation diagnostic. The associated proposal was classified
as bee in every frame, with confidence scores shown. All proposals were
drawn from the low-threshold tracking pool and remained below the final
output threshold of 0.01. The keyframe ground-truth annotation is
hoverfly.}
\label{fig:hoverfly-sequence}
\end{figure}

Figure~\ref{fig:hoverfly-confusions} shows selected, well-localized
bee--hoverfly classification failures, while
Figure~\ref{fig:hoverfly-sequence} shows a temporal failure case.

These results do not show that temporal information is unhelpful.
Rather, they identify a limitation of post-hoc processing applied after
independent frame-level classification. Score smoothing and trajectory
association can help when confidence is unstable or detections are
intermittent, but they cannot create correctly classified evidence when
neighboring frames contain none. A learned video detector,
optical-flow-aligned features, or feature-level temporal aggregation
could use motion and appearance jointly~\cite{fgfa,selsa,transvod}
before the final class decision and is therefore a more promising
direction.

\subsection{Limitations}
\label{sec:limitations}

The main limitation is that the final models were trained on the combined
training and validation splits, leaving no clean local holdout for unbiased
evaluation. Consequently, local results are used only for diagnosis, while
generalization is assessed through the limited number of hidden-test
submissions. The temporal methods were evaluated only on the contaminated former validation
split and, given the limited remaining submission budget, were not submitted to
the hidden-test server. Their negative result should therefore not be interpreted
as evidence about hidden-test generalization.

Each main training configuration was run once with a fixed seed because
of the available compute budget, so run-to-run variance was not measured
and small score differences should be interpreted cautiously. We also
did not record uniform measurements of inference latency, total training
time, and peak GPU memory across all systems. The study therefore does
not provide a complete accuracy--cost frontier.

The final phase also allowed only ten submissions, preventing hidden
evaluation of every trained configuration. The fresh 1536-pixel model
and the further-resumed 1536 model were not submitted, and their local
scores cannot establish how they would have generalized. Finally, the
temporal method used fixed post-hoc association, permitted no skipped
frames, and did not learn temporal features. Its results should
therefore not be generalized to end-to-end video detection methods.

\section{Conclusion}
\label{sec:conclusion}

We presented a compute-constrained empirical study of small-pollinator
detection in cluttered field video. RF-DETR outperformed the submitted
YOLO baselines, and increasing its input resolution from 1120 to 1344
produced our best hidden-test result of 0.405
$\mathrm{mAP}_{50:95}$. In contrast, the tested sliced-inference
fusion, size-routed ensemble, and warm-started 1536-pixel continuation
did not surpass the 1344-pixel model, while post-hoc temporal processing
did not improve the available local diagnostic evaluation. Error
analysis identified bee--hoverfly discrimination as the clearest
remaining bottleneck: neighboring frames rarely supplied correctly classified hoverfly evidence for post-hoc temporal correction. These results motivate temporal models that aggregate learned features across frames before the final classification decision.

\bibliographystyle{splncs04}
\bibliography{main}

\end{document}